# CAGD – Computer Aided Gripper Design for a Flexible Gripping System

**Michael Sdahl & Bernd Kuhlenkoetter**
University of Dortmund, Faculty of Mechanical Engeneering, Chair MGH, Germany
Michael.Sdahl@Uni-Dortmund.de
Bernd.Kuhlenkoetter@Uni-Dortmund.de

*Abstract: This paper is a summary of the recently accomplished research work on flexible gripping systems. The goal is to develop a gripper which can be used for a great amount of geometrically variant workpieces. The economic aspect is of particular importance during the whole development. The high flexibility of the gripper is obtained by three parallel used principles. These are human and computer based analysis of the gripping object as well as mechanical adaptation of the gripper to the object with the help of servo motors. The focus is on the gripping of free-form surfaces with suction cup.*
*Keywords: flexible gripper, CAD, suction gripper, flexibility in manufacturing.*

## 1. Current Situation

Grippers play an important role within automation systems. They are the interface between the workpiece and the whole automation system which realises a certain production process. Mostly a gripper design is a special and unique solution for a handling task of a given workpiece. Therefore the component "gripper" has an high impact on economical aspects, when flexible automation systems are needed (Seegräber, L., 1993).

In contrast to industrial robots, which can adapt themselves by programming to workpieces, most grippers in automated installations have to be changed depending on the workpiece. Economical solutions for flexible gripping solutions are rare.

At the moment the flexibility of industrial gripping systems is achieved by the use of revolver-grippers, gripper changing systems or gripper's jaw changing systems. Programmable grippers - the most expensive group of grippers - have the greatest flexibility compared to the others. They can adjust themselves to a workpiece with the help e.g. pneumatic or electric motors in combination with the use of sensors.

Several research projects exist among these industrial gripping systems. Researchers are working on the imitation of the flexibility of the human hand. Examples are the Karlsruher Hand (Osswald, D., 2001), the IFAS-Hand developed at RWTH Aachen (Czinki, A., 2001) and the DLR-Hand (Butterfass, J., 2001). They represent high-end technology not suitable for general industrial use.

There are some applications where adaptive industrial gripper solutions could be helpful to decrease setup times of handling systems and achieve a high flexibility of the whole automation system. One example deals with the handling of car reflectors. Automobile suppliers have to produce different kinds of car reflectors, sometimes hundreds of versions for current car types and for the accessories market. Especially quality control systems have to deal with all those kinds of reflectors and therefore need a high flexible gripping system to handle the reflector types. For this task an adjustable, adaptive gripping system could be an economical improvement.

First developments on flexible gripping technology are carried out in this article.

## 2. Flexible gripper development

This article mainly focuses on a three finger gripper with suction cups as a possible solution for a flexible gripper design. In order to develop a flexible gripping system for industrial use first of all the following three topics are analysed:

- Human based analysis of gripping objects
- Computer based analysis of gripping objects and
- Mechanical adaptation of the gripper to gripping objects

The first two principles are used to find uniform gripping point constellations on similar but geometrically different workpieces. If this is possible no adaption of the gripper to the workpiece is necessary what would safe set-up time.

To grip several hundred workpieces of varying geometry it can be assumed, that there is the need for different constellations of adjustable (adaptive) gripping arms. Of course, the number of different gripper configurations should be as low as possible.

The human workpiece analysis is useful for a small number of objects, when used in context with the goal of finding uniform gripping constellations. As the amount of workpieces for analysis increase, the time and effort for the examination rises exponentially. This is the starting point for computer based analysis, which analyses CAD data and computes similar gripping points on different workpieces.

*2.1 Human workpiece analysis as a basis for the design of a gripper*
The first step in designing a gripper for different workpieces is to identify geometrical similarities of the workpieces. This is difficult for a human, because one has to detect the same gripping points on different workpieces. The most common method in such a case is "Trial and Error". The gripping surface found on the first gripping object and the associated gripping constellation is transferred to the next gripping object. As soon as the second workpiece can be gripped, the iteration process is finished, as long as only two workpieces need to be analysed. In the most cases, it is not possible to transfer one gripping constellation from the first workpiece to the second one (in cases of highly geometrically varying workpieces). Hence, several iteration steps are necessary before a constellation can be found, which fits to two different workpieces. The number of iteration loops increases notably, if more than two workpieces are involved.

**3. Design of a gripper based on a CAD data analysis – CAGD**

To solve the problem of designing a gripper for a huge amount of workpieces, the use of software is helpful. This software should detect possible gripping points on different CAD workpieces. The goal is to find similar gripping areas on different workpieces, so that a gripper can be designed with the least amount of adjusting mechanisms or, if possible no moving parts.

*3.1 Problem definition and description of the to be solved task*
The comparison with human procedure helps to identify the problem, which has to be solved. When a human is determining gripping points for suction grippers, one identifies the gripping points by mostly viewing the workpiece. Boundary conditions are considered more intuitively, such as accessibility to the gripping point, surface's condition (e.g. roughness), centre of gravity and other dimensions of the workpiece as well as parameters, which result from the following process.

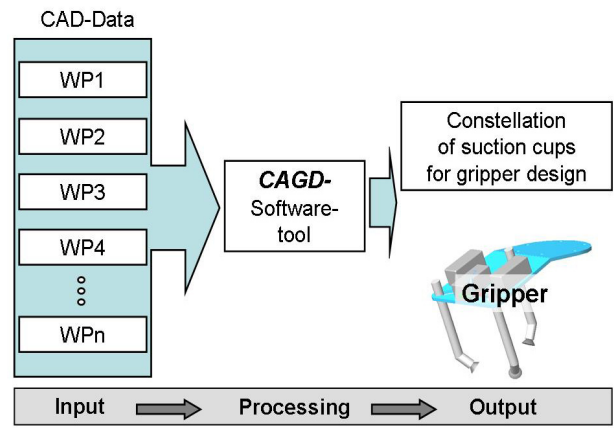

Fig. 1. IPO-principle in CAGD

In order to get first results in case of a huge variety of workpieces an easy to use algorithm was developed which is based on the CAD-Data and a CAD-Kernel (Parametric Technology Corporation, 2004; Hoschek, J., 1989). To find gripping points with the help of the software, information about the surface of the workpiece is necessary. This information is presented in the form of CAD data. The CAD drawings of all the to-be-gripped workpieces are loaded successively into the software. The software detects gripping points on the workpieces depending on given boundary conditions. The result will be a constellation of gripping areas to grip all the analysed workpieces via a suction gripper (shown in Fig. 1). If it is possible, the result will be one constellation. Also, gripping constellation of a special pattern can be found. This pattern describes the workspace of a flexible gripper, see Fig. 4.

In most cases, the same constellation cannot be used on different workpieces. Therefore, tolerances have to be given for the examination. The tolerances refer to different curvature and angles to the horizontal as well as height and transverse deviations toward the normal vector and/or orthogonal to the direction of the normal vector of the suction area.

The boundary conditions for the gripping points are passed to the software via an input mask. At the current level of development the parameters are: diameter of the suction cup, minimum distance of the suction cups to each other, as well as further data concerning the necessary surface conditions of the suction face (e.g. roughness and curvature).

*3.2 Gripping point determination with n workpieces*
The necessary information about the workpieces is provided by CAD data. With these data combined with the use of the existing functions in the CAD kernel (Parametric Technology Corporation, 2004 ; Hoschek, J. 1989) different computations are executed for gripping point determination.

A help geometry (cylinder 1 shown in Fig. 2) with the same diameter as the vacuum cup is necessary for the determination of the gripping point. This one is cut with the CAD workpiece and can contain geometrical boundary conditions, which are made to the gripping area. In the simplest case the help geometry corresponds - as shown - to a cylinder, which is intersected with the surface of the to be examined workpiece in normal direction.

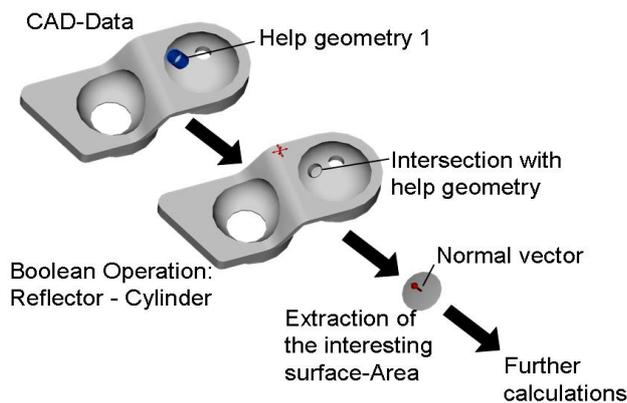

Fig. 2. Extraction of the to be examined surface area

The normal vectors are of great importance for the following computations. A characteristic comparison of all normal vectors on the surface of the CAD workpiece makes it possible to limit the number of to-be-examined surfaces from the beginning. In this manner, the expense of computation is reduced dramatically.

The procedure for determining the first gripping area can be described as follows: The CAD data and the help geometry are loaded into the software. The workpiece is then embedded in a three-dimensional raster before the gripping point determination. Thus, it is possible to have a finite quantity of to-be-examined points. The result of the cuts on the workpiece and the raster are points on the surface of the workpiece, which are starting points for the cut on the CAD workpiece with the help geometry. Now the help geometry is aligned on the basis of the normal vector. In the simple example shown, the cylinder's symmetry axis would have the direction of the normal vector.

The cylinder cuts a segment out of the surface of the CAD workpiece. This function can be found, for example, in all currently available CAD programs (subtraction of bodies).

On the surface segment, different computations and characteristic inquiries are run with the help of the CAD-Kernel to be able to make the decision whether it is a useful gripping point for a suction gripper or not. The representation of the individual computational steps with mathematical and kernel specific commands are not included in this paper.

Fig. 3 shows the procedure for the determination of the next gripping point. The next gripping point should be a minimum distance from the first gripping point. This condition is fulfilled with the use of another help geometry. A second cylinder is intersected with the surface of the CAD workpiece (cylinder 2 in *Fig. 3*).

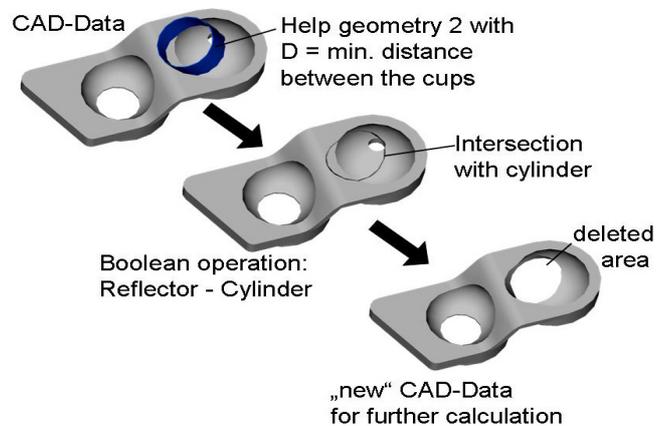

Fig. 3. Generation of new CAD-Data

This cylinder intersects exactly at the point, at which the first cylinder intersects the workpiece and where the first gripping point had been determined. The radius of this cylinder corresponds to the minimum distance, which the next gripping point must have to the first gripping point. The intersection plane (CAD workpiece - cylinder) is cut out of the CAD workpiece. The resulting CAD workpiece is stored as a new workpiece and is used as input for determining the next gripping point. Using this method, it is ensured that only gripping points are found in a defined minimum distance.

If a gripper constellation for a workpiece is found, it is tested on the next workpiece. If the second one passes the test, the third workpiece is loaded and the gripper configuration is again tested. However, the second workpiece will most likely fail the test on its first attempt, unless the geometry is very similar to the first workpiece. In this case, the first step is repeated and another gripper configuration is searched on the first workpiece. The process is repeated till a similar gripper configuration has been found to grip all the workpieces loaded in the CAGD-Software.

The requirements on gripping areas for suction grippers are partly realized by the help geometry or by computations and comparison of values with the help of the CAD-kernel. The more boundary conditions in the help geometry are considered the faster is the computation. If one of the following conditions does not apply, the computation is stopped and runs again at a new point:

- *Gripping faces must not exhibit unevenness*

  The surfaces, on which the vacuum cups touch down, may exhibit only small unevenness. The tolerance of the vacuum cup to unevenness can be given to the system before calculation. The easiest way is with the use of another help geometry. A cone is a good defining element, which can be symmetrically located on a spherical segment of the analysed CAD workpiece. The centre of this geometry is aligned with the cut surface section of the workpiece. The symmetry axis of this help geometry points toward the normal vector which is in the centre of the surface section. The diameter is identical to that of the vacuum cup. As soon as this help geometry cuts the cut surface element, it is an indication that the vacuum cup cannot apply. The computations are stopped and a new surface element is examined.

- *Gripping areas must have a defined distance to each other*

The vacuum cups should be set evenly around the centre of mass of the workpiece. This is obtained with the help of the defined second help geometry as shown in Fig. 3, which excludes that suction faces are found in direct proximity to one another.

- *Gripping points must not be on a line*

If this would be the case, the workpiece would tip over. To prevent this, the furthest points away are connected by a line and the distance between this line and a third point is determined. If the distance is below a given minimum, the computation is stopped and a new gripping point is determined.

### 4. Mechanical adjustment

The last possibility, specified in this paper, for increasing the flexibility of a gripping system consists of assembling the suction cups to an adjusting mechanism. With this, it is possible to adapt the vacuum cups to different workpieces.
A naive beginning would be to copy the human hand and assemble vacuum cups at the fingers. This gripper would be able to adapt itself to nearly every workpiece and grip it with vacuum cups. This enormous flexibility of the hand is anatomically given by the high number of hand bones (27) and a complex system of muscles and tendons, thereby allowing the human hand to have 22 degrees of freedom (Fuller, J. L., 1991).
It is assumed that the workpieces are within certain limits of physical boundary conditions (dimensions and weight). The development and production of a gripper based on the mechanical system of a human hand is expensive. Thus, only the necessary number of degrees of freedom is included. The drive of such a system must have a low frequency of maintenance, be cheap but reliable, and simple to control in order to provide quick industrial implementation. A PLC is used in this case to change the gripper constellation. A signal to adjust the gripper can come from a bar code, magnetic stripe or an image processing system.

### 5. Combination of the different possibilities

By combining the different alternatives represented in this paper it is possible to simplify the adaptation mechanisms (Fig. 4), because several gripping points can be found by the CAGD-software. Consequently, the gripper can be equipped with a minimum amount of degrees of freedom. In dependence of the workpiece examined by the software a gripper constellation for all workpieces can be found. This makes the use of an adjustment mechanism redundant.

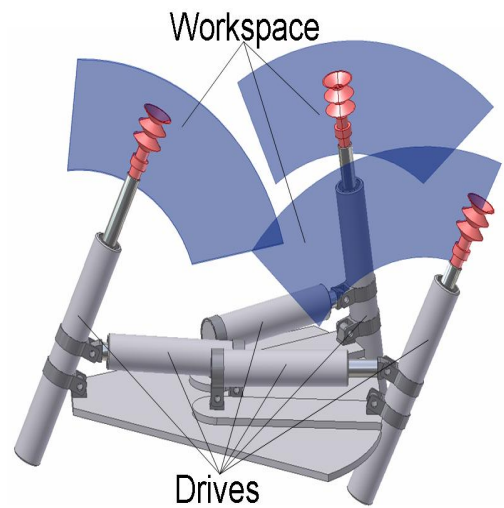

Fig. 4. Possible arrangement of drives and the resulting workspace

### 6. Perspective

Presently, the goal is to link the system with a 3D-image processing system. The data from the image processing system will be loaded into the software presented in this paper. Hence, gripping points will be handed over automatically to the gripping system. For this, a laser scanner could also be used, which provides a 3D-picture of the workpiece. Therefore, it could be possible to grip workpieces automatically in the future.
Moreover after the prototype phase the software with the algorithms to find the gripping points has to be improved. Intelligent mathematical algorithms will improve computing time and optimise the gripping constellation. Contemporaneous more boundary conditions can be taken into account with the needed accuracy.